\documentclass[sigconf]{acmart}

\AtBeginDocument{%
  }

\copyrightyear{2023}
\acmYear{2023}
\setcopyright{acmlicensed}\acmConference[KDD '23]{Proceedings of the 29th ACM SIGKDD Conference on Knowledge Discovery and Data Mining}{August 6--10, 2023}{Long Beach, CA, USA}
\acmBooktitle{Proceedings of the 29th ACM SIGKDD Conference on Knowledge Discovery and Data Mining (KDD '23), August 6--10, 2023, Long Beach, CA, USA}
\acmPrice{15.00}
\acmDOI{10.1145/3580305.3599906}
\acmISBN{979-8-4007-0103-0/23/08}

\usepackage{multirow}
\usepackage{balance}

\begin{document}

\title{ShuttleSet: A Human-Annotated Stroke-Level Singles Dataset for Badminton Tactical Analysis}


\author{Wei-Yao Wang}
\orcid{0000-0002-6551-1720}
\affiliation{%
  \institution{National Yang Ming Chiao Tung University}
  \city{Hsinchu}
  \country{Taiwan}}
\email{sf1638.cs05@nctu.edu.tw}

\author{Yung-Chang Huang}
\orcid{0009-0006-6648-0598}
\affiliation{%
  \institution{National Yang Ming Chiao Tung University}
  \city{Hsinchu}
  \country{Taiwan}}
\email{jason880102.cs10@nycu.edu.tw}

\author{Tsi-Ui Ik}
\orcid{0000-0001-6432-9161}
\affiliation{%
  \institution{National Yang Ming Chiao Tung University}
  \city{Hsinchu}
  \country{Taiwan}}
\email{tik@nycu.edu.tw}

\author{Wen-Chih Peng}
\orcid{0000-0002-0172-7311}
\affiliation{%
  \institution{National Yang Ming Chiao Tung University}
  \city{Hsinchu}
  \country{Taiwan}}
\email{wcpengcs@nycu.edu.tw}

% \author{Wei-Yao Wang, Yung-Chang Huang, Tsi-Ui Ik, Wen-Chih Peng}
% \affiliation{%
%   \institution{National Yang Ming Chiao Tung University}
%   \city{Hsinchu}
%   \country{Taiwan}}
% \email{sf1638.cs05@nctu.edu.tw, jason880102.cs10@nycu.edu.tw, cwyi@nctu.edu.tw, wcpeng@cs.nycu.edu.tw}

\renewcommand{\shortauthors}{Wei-Yao Wang, Yung-Chang Huang, Tsi-Ui Ik, \& Wen-Chih Peng}

\begin{abstract}
With the recent progress in sports analytics, deep learning approaches have demonstrated the effectiveness of mining insights into players' tactics for improving performance quality and fan engagement.
This is attributed to the availability of public ground-truth datasets.
While there are a few available datasets for turn-based sports for action detection, these datasets severely lack structured source data and stroke-level records since these require high-cost labeling efforts from domain experts and are hard to detect using automatic techniques.
Consequently, the development of artificial intelligence approaches is significantly hindered when existing models are applied to more challenging structured turn-based sequences.
In this paper, we present ShuttleSet, the largest publicly-available badminton singles dataset with annotated stroke-level records.
It contains 104 sets, 3,685 rallies, and 36,492 strokes in 44 matches between 2018 and 2021 with 27 top-ranking men's singles and women's singles players.
ShuttleSet is manually annotated with a computer-aided labeling tool to increase the labeling efficiency and effectiveness of selecting the shot type with a choice of 18 distinct classes, the corresponding hitting locations, and the locations of both players at each stroke.
In the experiments, we provide multiple benchmarks (i.e., stroke influence, stroke forecasting, and movement forecasting) with baselines to illustrate the practicability of using ShuttleSet for turn-based analytics, which is expected to stimulate both academic and sports communities.
Over the past two years, a visualization platform has been deployed to illustrate the variability of analysis cases from ShuttleSet for coaches to delve into players' tactical preferences with human-interactive interfaces, which was also used by national badminton teams during multiple international high-ranking matches.
\end{abstract}




\maketitle

\section{Introduction}

%% recent progress of sports analytics
%% 帶application
%% 帶data from other sports
%% 帶羽球沒有
With the advancement of artificial intelligence in recent years, sports analytics has significantly changed the sports environment with video understanding and match analysis using machine learning (ML) and deep learning (DL) approaches.
For example, a video assistant referee system has been deployed in football over 100 competitions including the 2022 FIFA World Cup to support the referee team to reduce misjudgments\footnote{https://www.fifa.com/technical/football-technology/football-technologies-and-innovations-at-the-fifa-world-cup-2022/video-assistant-referee-var}.
On the other hand, player evaluation and tactic investigation have improved the effectiveness in multiple sports, such as football \cite{DBLP:conf/kdd/DecroosBHD19,DBLP:conf/aaai/AzadKWLSASS22,DBLP:conf/kdd/KimKCYK22,DBLP:conf/kdd/SimpsonBLN22}, baseball \cite{DBLP:conf/hci/KincaidGJTKB21,DBLP:journals/entropy/HanKNK22}, basketball \cite{DBLP:conf/cikm/ChenJJZLB022,DBLP:conf/aaai/YanaiSKSR22}, and badminton \cite{DBLP:conf/aaai/WangSCP22,DBLP:journals/corr/abs-2211-12217}.
To train models for these applications, the datasets are mainly collected from either videos \cite{DBLP:conf/cvpr/DeliegeCGSDNGMD21,DBLP:conf/cvpr/ZhuW022} or sensors \cite{DBLP:conf/momm/ItoG18,DBLP:journals/tiot/HaladjianSTTB20,DBLP:journals/sensors/PillitteriTBNSG21} to extract the enormous variable metadata in the matches.
Therefore, it is beneficial for researchers to conduct analysis and introduce fruitful findings from the collected dataset that are able to foster not only coaches and players but also sports media and spectators.

\begin{figure}
    \centering
    \includegraphics[width=\linewidth]{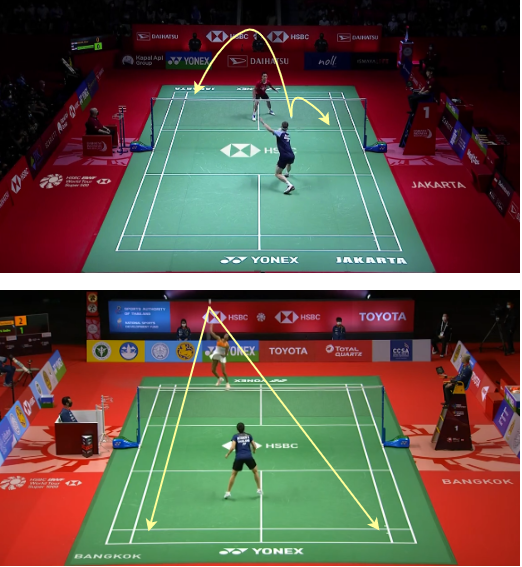}
    \caption{Two examples with potential returning strokes of men's and women's singles.}
\end{figure}

%% public datasets in other sports
Generally, there are several available datasets with frame-based or event-based records for various sports; for instance, \citet{DBLP:conf/cvpr/DeliegeCGSDNGMD21} proposed SoccerNet-v2 for soccer understanding and broadcast production via annotated soccer match videos, and \citet{DBLP:conf/icdm/YueLCBM14} released SportsVU, which is a tracking dataset for games recorded from several seasons of NBA.
However, in turn-based sports, there are only a few public datasets with tournament-level records \cite{tennis_dataset1,tennis_dataset2}, but none of them are labeled with the stroke-level records due to the human cost and domain gap for labeling (e.g., what shot type a player performs).
These issues hinder researchers' efforts to design techniques to analyze demanding applications such as hitting patterns, tactic investigations, and action recognition from such turn-based sequences \cite{DBLP:conf/cikm/Wang22}.
As illustrated in Figure 1, players have multiple choices to organize their tactics so as not to be seen through by their opponents in badminton matches, and thus pre-match data collection and pattern investigation are needed to forecast the next stroke on the court.
% the complex tactics in badminton matches make players unpredictable for not being seen through with only the instant circumstances and thus need pre-match data collection and pattern investigation to forecast the next stroke on the field.
Therefore, we aim to create microscopic metadata from badminton matches to accurately describe the process from strokes to matches.

%% challenge
Although existing automatic detecting tools, e.g., \cite{DBLP:conf/mir/ChuS17}, can be adopted to extract spatial-temporal stroke features from badminton videos, the automation introduces a constraint on the variability in the dataset since corresponding structured source data (i.e., the compositions of stroke, rally, set, and match) must be available to analyze the similarities and discrepancies between players within a match as well as across matches.
Furthermore, providing high-quality annotation requires serious efforts from domain experts since not only the number of videos is rapidly increasing but it is also hard for non-domain experts to distinguish the tactical metadata of a stroke.
% but also the tactical metadata of a stroke is often hard to be distinguished by non-domain experts.
Therefore, one of the key challenges is to implement a labeling tool that can provide a straightforward process for labeling from each stroke to each match to reduce the difficulty of annotators reviewing videos repeatedly.

%% Our dataset
In this paper, we present the ShuttleSet dataset annotated by domain experts with the efficient and precise S\textsuperscript{2}-labeling tool \cite{DBLP:conf/apnoms/HuangHLIW22} using the BLSR format \cite{10.1145/3551391}.
Based on the labeling tool, ShuttleSet provides stroke-by-stroke human annotation ground truth records for 44 matches from 2018 to 2021 with 27 high-ranking players (16 men's singles and 11 women's singles), which consist of 104 sets, 3,685 rallies, and 36,492 strokes.
Following the current public sports dataset, ShuttleSet will likewise be made available to the public in order to stimulate both the research and sports analysis communities for more advanced applications.

%% Summarized contributions
Our ShuttleSet dataset distinguishes itself in the following aspects:
\begin{enumerate}
    \item[1)] \textbf{The Largest Public Stroke-Level Dataset:} To the best of our knowledge, this is the largest public turn-based sports dataset with stroke-level records which attempts to mitigate the gap between turn-based sports and research communities.
    \item[2)] \textbf{Efficient Expert Annotation:} In contrast to automation approaches to generate the data set, we rely on annotations from domain experts to accurately describe the detailed and structured process during a badminton match. Moreover, we adopt a computer-aided labeling tool with user-friendly designs and computer vision techniques to reduce the burden of watching the video repeatedly.
    \item[3)] \textbf{Detailed Fields:} The fields in the dataset include temporal, spatial, posture, and skill categories, which can be used for spatial-temporal mining, action recognition, and tactic investigation.
    \item[4)] \textbf{Customized Train, Validation, and Test Sets:} As we do not intend to limit the scope of the application usage (i.e., no fixed labels), ShuttleSet provides the customized availability to split the datasets instead of the fixed train-, validation, and test-sets.
    \item[5)] \textbf{Paradigms of Practical Usage:}
    We illustrate multiple usages in Sections \ref{experiments} and \ref{visualization-platform} and provide a Python implementation of the ShuttleSet dataset\footnote{https://github.com/wywyWang/CoachAI-Projects.} to advance the research of multifaceted applications.
\end{enumerate}
\section{Related Work}
Nowadays, with the development of video understanding technologies, sports analytics has improved player and team performance and raised social engagement by mining effective insights to address several challenges between sports analytics and artificial intelligence \cite{DBLP:conf/ijcai/DecroosBHD20}.
To collect previous match records for analyzing player behaviors, researchers have designed applications for detecting actions from videos automatically \cite{DBLP:journals/corr/abs-2206-01038}, e.g., Sports-1M \cite{DBLP:conf/cvpr/KarpathyTSLSF14}, UCF Sports \cite{DBLP:conf/cvpr/RodriguezAS08}, and J-HMDB \cite{DBLP:conf/iccv/JhuangGZSB13}.
Recently, \citet{DBLP:conf/iccv/LiCH0W021} presented a multi-person dataset consisting of spatio-temporal localized sports actions, which includes faster movement and multiple concurrent actions by multiple players.
SoccerNet-v2 is collected to encourage future research to advance soccer understanding and broadcast production with action recognition and camera shot segmentation \cite{DBLP:conf/cvpr/DeliegeCGSDNGMD21}.
These public datasets motivate the research community to develop the research of these sports and understand teams' and individuals' behaviors in fulfilling the coaches' strategies.

\begin{figure}
    \centering
    \includegraphics[width=\linewidth]{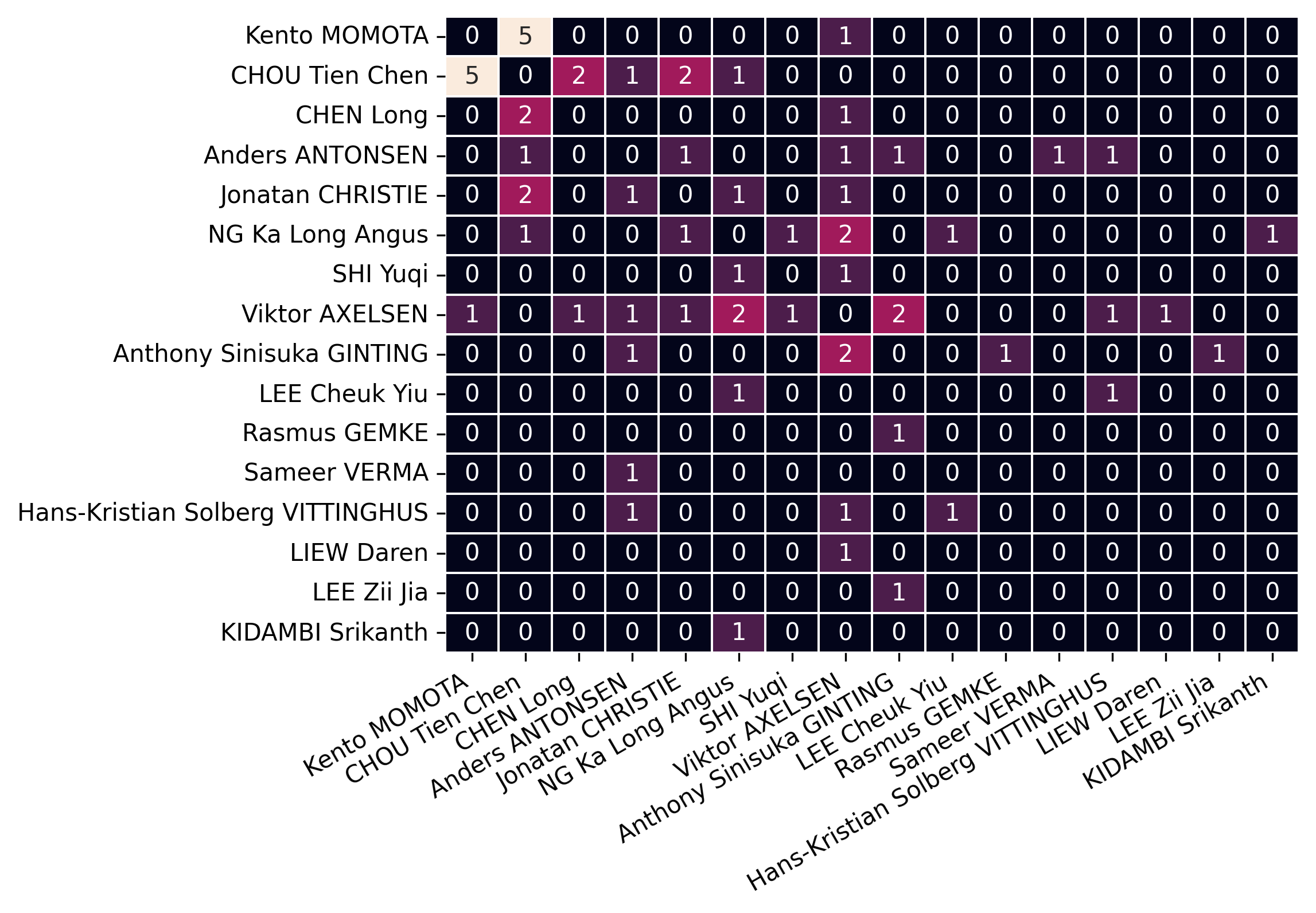}
    \caption{The matchup matrix of men's singles.}
    \label{men-matrix}
\end{figure}

In the field of badminton, existing approaches mainly focus on action and stroke detection from videos as well.
\citet{DBLP:conf/mir/ChuS17} developed an interface illustrating court detection, shot type classification, and offensive/defensive classification results from the given video.
\citet{Yoshikawa2021} incorporates skeleton information to detect the timing of the overhead stroke by their collected videos with specialized cameras.
To broaden the availability of videos to the general matches, an automatic annotation framework was designed to detect players from the official broadcast videos without special cameras or additional sensors \cite{DBLP:conf/wacv/GhoshSJ18}.
Although these techniques are useful for tracking the motion recognition of players to construct datasets automatically \cite{DBLP:journals/fgcs/FangS21}, it remains a constraint of the opportunity of structured data from source videos, which obstructs researchers from conducting various analyses for delving into advanced tactic investigations.
In addition, there is no existing public turn-based dataset with sufficient stroke-level records, which requires high-quality annotation from domain experts for labeling.
Therefore, we propose ShuttleSet as the largest stroke-level badminton dataset for turn-based sports analytics.
The aim is to not only foster the academic community's research on turn-based sequences but also the sports community's development of applications to enhance their tactical strategies.
\section{The ShuttleSet Dataset}
\label{shuttleset-dataset}
\subsection{Overview}
ShuttleSet contains 36,492 strokes, 3,685 rallies, and 104 sets from 44 matches between 2018 and 2021 played by 27 high-ranking men's and women's players.
The details of annotation matches and the corresponding players are listed in Appendix \ref{detail-match}.
Each match consists of 2 or 3 sets (best of 3), each of which is comprised of the metadata of strokes to describe the microscopic process from stroke to rally.
The format of each stroke can be divided into four categories, which mainly expands from the BLSR design format \cite{DBLP:conf/icdm/WangCYWFP21}:
\begin{itemize}
    \item Rally category: 1) The current scores of both players in the set. 2) The player who won the rally, which only appears in the last stroke of each rally. 3) Lose reason is classified from \textit{out}, \textit{touched the net}, \textit{not pass over the net}, \textit{opponent's ball landed}, and \textit{misjudged}.
    \item Temporal category: 1) The hitting time of the stroke. 2) The frame of the hitting stroke. We note that the frame number can be converted to the hitting time based on the frame rate of the video.
    \item Spatial category: 1) The locations of both players on the court at the hitting time. 2) The hitting locations of the shuttle at the hitting time. These locations are recorded by both accurate x and y coordinates and overall grids. 3) The height at the hitting time, which is a binary variable indicates whether the hitting height is below or above the net. This field can be used to determine if the player returns a stroke passively or aggressively.
    \item Hitting category: 1) The status of the hitting stroke, which indicates if the stroke belongs to the serve, return, or dead bird type. 2) Hit the shuttle with the backhand or not. 3) Hit the shuttle around the head or not. 4) The shot type of the hitting stroke, where the shot type is selected from one of the 18 types: \textit{net shot}, \textit{return net}, \textit{smash}, \textit{wrist smash}, \textit{lob}, \textit{defensive return lob}, \textit{clear}, \textit{drive}, \textit{driven flight}, \textit{back-court drive}, \textit{drop}, \textit{passive drop}, \textit{push}, \textit{rush}, \textit{defensive return drive}, \textit{cross-court net shot}, \textit{short service}, and \textit{long service}. The detailed explanations of each type are introduced in \cite{10.1145/3551391}.
\end{itemize}

\begin{figure}
    \centering
    \includegraphics[width=\linewidth]{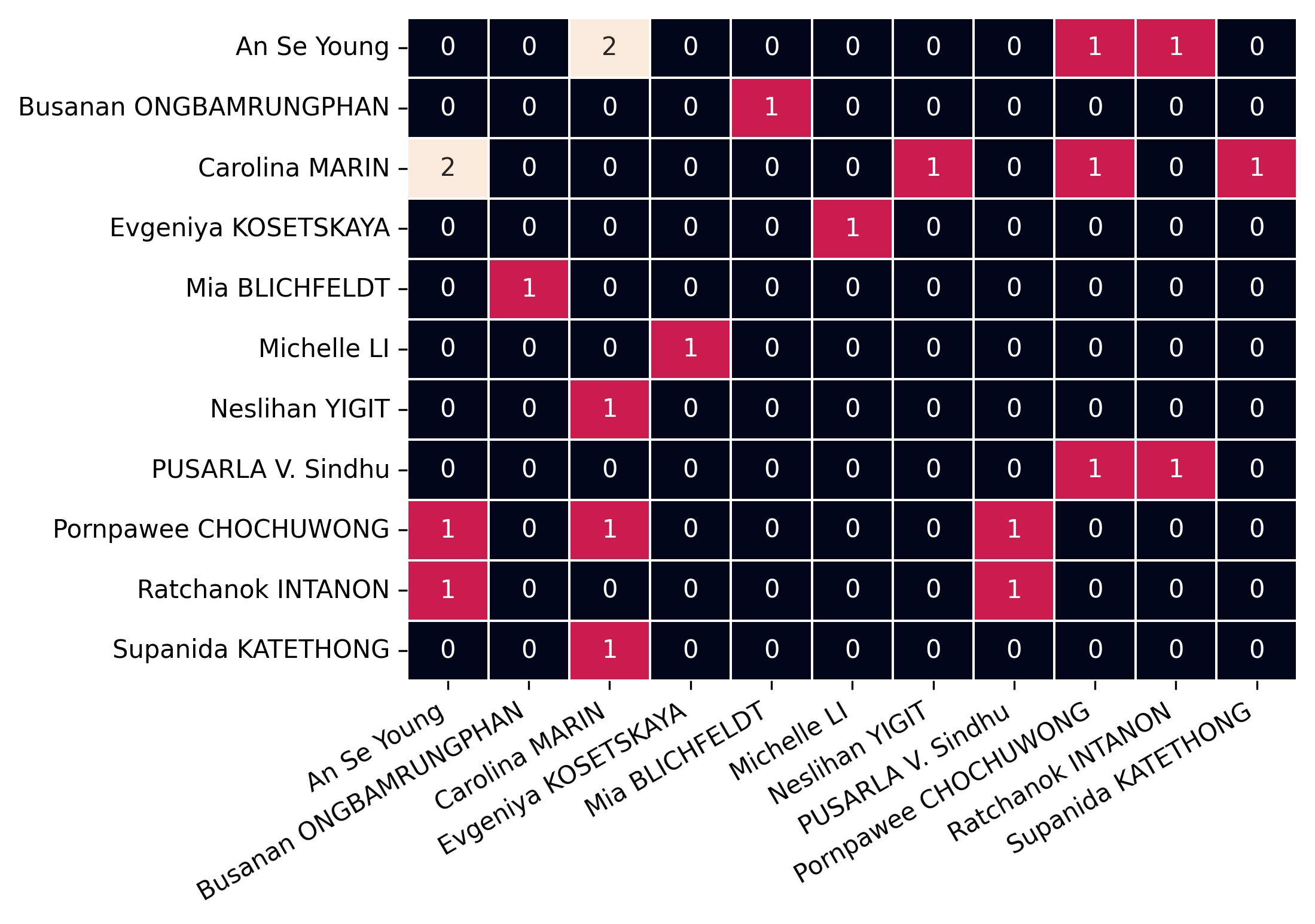}
    \caption{The matchup matrix of women's singles.}
    \label{women-matrix}
\end{figure}

In addition to the definition of stroke-level records, we require that the labeled matches in ShuttleSet adhere to a few conditions.
First, the matches are selected from quarter-finals, semi-finals, and finals\footnote{There are a few matches are from the group stage of BWF World Tour Finals.} of top-selective tournaments (e.g., Super 1000 series, BWF World Tour Finals) to provide state-of-the-art and highly-organized tactics from top players, which can be used to investigate players' tactics and improve the hitting quality of low-ranking and junior players.
As a second condition, videos are crawled from public sources\footnote{http://bwf.tv/}, which not only broadens the available usage of videos to general matches but also does not require specialized cameras to collect videos.
Third, ShuttleSet consists of both men's and women's singles matches to obtain maximum variability, which can be analyzed for the different hitting patterns from different matchups.
Moreover, the number of players and the number of the played count of each player are various in ShuttleSet to enable researchers to investigate the hitting patterns from the same player and between different players, which also points out the challenging applications due to the rare-appearance players.
In Figures \ref{men-matrix} and \ref{women-matrix}, we demonstrate the matching matrix of each player contained in ShuttleSet with the corresponding appearances.

\begin{figure}
    \centering
    \includegraphics[width=\linewidth]{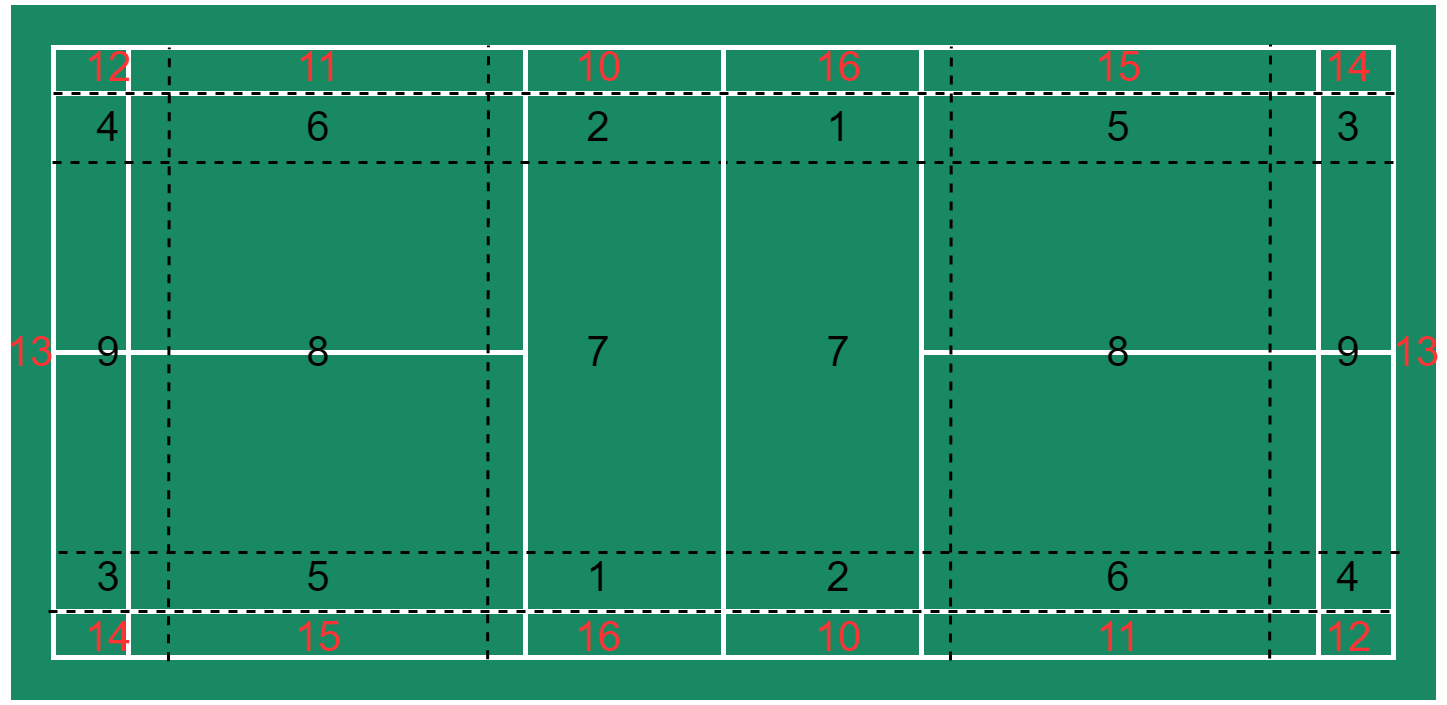}
    \caption{The grid system layout on the badminton court. Numbers denote the corresponding locations. The red color is outside the singles court.}
    \label{court}
\end{figure}

The BLSR format only adopts the grids on the badminton court to record location systems for traditional analysis (e.g., \cite{valldecabres2020players,Wang2020badminton}), which fails to record the fine-grained coordinates of both players and shuttles.
To integrate the traditional tactical analysis, the conversion from 2D coordinates of players and shuttles is designed to map them to the corresponding grids as illustrated in Figure \ref{court}.
In general, the grid system is introduced by domain experts with nine grids for locations in the court and an additional seven grids for locations outside the court.
As the court is divided into two symmetric sub-courts, the grids of each sub-court are identical to provide a consistent format for location analysis.
Formally, the conversion can be achieved by first converting the camera coordinate system to the real-world coordinate system:
\begin{equation}
    p' = Hp,
\end{equation}
where $p'$ is the 2D real-world coordinate system, $p$ is the 2D camera coordinate system, and $H \in \mathbb{R}^{3 \times 3}$ is a homography matrix between these systems, which can be solved by providing at least four pairs between two coordinate systems.
The corners of the court (i.e., (0, 0), (0, 6.1), (13.4, 0), (13.4, 6.1)) are consistent since the court in the real world is 13.4 meters in length and 6.1 meters in width; therefore, the corners' coordinates in the camera coordinate system can be acquired by only labeling them in the camera coordinate system.

%% sample matches consistency
As ShuttleSet is annotated by domain experts with a computer-aided labeling tool, we sample three matches for labeling again to ensure data consistency.
For the temporal category, the time differences of serve, return, and dead bird are 0.24, 0.02, and 0 frames in terms of the mean absolute error (MAE), respectively.
For the spatial category, the MAE of the shuttle's position is 0.18 on average and the MAE of the player's position is 0.86 on average.
For the labels of shot type, the quality difference of serve type, return type, and dead bird type is 0, 0.05, and 0.11 in terms of accuracy.

Table \ref{type-distribution} summarizes the distribution of each shot type of both men's and women's singles in ShuttleSet.
The two most common shot types of men's and women's singles are \textit{net shot} and \textit{lob}.
The third most common type of men's singles is \textit{return net} and that of women's singles is \textit{clear}.
This also boosts the variability of the dataset by mixing stroke-level records of both men's and women's singles.
In addition, ShuttleSet contains different proportions of shot types, which might be a challenge for designing turn-based methods to take imbalance types into account.

Since the goal of ShuttleSet is to provide turn-based strokes to stimulate advanced research on turn-based sports, we do not split up our dataset into pre-defined train-, validation-, and test-sets.
In contrast, we provide multiple real-world but challenging analysis paradigms in the experiments to showcase the usage of ShuttleSet, which can also be used as the benchmarks for the research community for developing new algorithms. 

\begin{table}
    \small
    \centering
    \caption{Distribution of the shot type in ShuttleSet. Along with the frequency of each class label, we present the relative occurrence based on men's and women's singles.}
    \begin{tabular}{c|c|c|c|c}
    \toprule
    Shot type & Singles & Count & Proportion (\%) & Total (\%) \\
    \midrule
         \multirow{2}{*}{net shot} & Men & 3,781 & 15.45 & \multirow{2}{*}{17.60} \\
         & Women & 527 & 2.15 & \\
    \midrule
         \multirow{2}{*}{return net} & Men & 2,130 & 8.70 & \multirow{2}{*}{10.13} \\
         & Women & 349 & 1.43 & \\
    \midrule
         \multirow{2}{*}{smash} & Men & 1,430 & 5.85 & \multirow{2}{*}{7.13} \\
         & Women & 314 & 1.28 & \\
    \midrule
         \multirow{2}{*}{wrist smash} & Men & 950 & 3.88 & \multirow{2}{*}{4.61} \\
         & Women & 178 & 0.73 & \\
    \midrule
         \multirow{2}{*}{lob} & Men & 3,124 & 12.76 & \multirow{2}{*}{14.68} \\
         & Women & 470 & 1.92 & \\
    \midrule
         \multirow{2}{*}{defensive return lob} & Men & 149 & 0.61 & \multirow{2}{*}{0.76} \\
         & Women & 36 & 0.15 & \\
    \midrule
         \multirow{2}{*}{clear} & Men & 1,417 & 5.79 & \multirow{2}{*}{7.51} \\
         & Women & 421 & 1.72 & \\
    \midrule
         \multirow{2}{*}{drive} & Men & 451 & 1.84 & \multirow{2}{*}{2.06} \\
         & Women & 52 & 0.22 & \\
    \midrule
         \multirow{2}{*}{driven flight} & Men & 43 & 0.18 & \multirow{2}{*}{0.18} \\
         & Women & 0 & 0.00 & \\
    \midrule
         \multirow{2}{*}{back-court drive} & Men & 274 & 1.12 & \multirow{2}{*}{1.36} \\
         & Women & 59 & 0.24 & \\
    \midrule
         \multirow{2}{*}{drop} & Men & 1,112 & 4.54 & \multirow{2}{*}{5.68} \\
         & Women & 279 & 1.14 & \\
    \midrule
         \multirow{2}{*}{passive drop} & Men & 811 & 3.31 & \multirow{2}{*}{3.73} \\
         & Women & 101 & 0.42 & \\
    \midrule
         \multirow{2}{*}{push} & Men & 1,736 & 7.09 & \multirow{2}{*}{8.51} \\
         & Women & 347 & 1.42 & \\
    \midrule
         \multirow{2}{*}{rush} & Men & 296 & 1.21 & \multirow{2}{*}{1.36} \\
         & Women & 37 & 0.15 & \\
    \midrule
         \multirow{2}{*}{defensive return drive} & Men & 212 & 0.87 & \multirow{2}{*}{1.10} \\
         & Women & 56 & 0.23 & \\
    \midrule
         \multirow{2}{*}{cross-court net shot} & Men & 843 & 3.44 & \multirow{2}{*}{3.99} \\
         & Women & 135 & 0.55 & \\
    \midrule
         \multirow{2}{*}{short service} & Men & 1,728 & 7.06 & \multirow{2}{*}{8.16} \\
         & Women & 271 & 1.10 & \\
    \midrule
         \multirow{2}{*}{long service} & Men & 183 & 0.75 & \multirow{2}{*}{1.45} \\
         & Women & 172 & 0.70 & \\
    \bottomrule
\end{tabular}
    \label{type-distribution}
\end{table}

Despite being cost-intense and far less scalable than automatic labeling tools, the human annotation has several benefits over automated ground-truth generation.
The first and most obvious reason to adopt human annotations is the freedom to annotate any type of match without customizing a programmatic source.
For most matches, the camera angles are different due to the stadium's condition, and the spectators, referees, and sometimes matches from other courts are also filmed in the video.
These issues are not a hard constraint with human annotation since it is simple to capture the labeling target (e.g., players and shuttles) from the video, while these issues introduce noise for automatic methods.
A second reason to leverage human annotation is that it usually supports a more natural and structural interpretation of describing the process from a match to a rally, which requires severe efforts for both.
Automatic techniques on turn-based sports need not only detection accuracy but also complex processing to label end-to-end (e.g., detecting if the current set is an end).
This might be challenging even for human annotation; we, therefore, incorporate a computer-aided labeling tool to simplify the annotation process as outlined in Section \ref{annotation-process}.
Third, labeling ground truth through human annotation is usually considered a natural upper bound on the dataset quality, which is able to guarantee the performance expectation in training deep neural network models and avoiding overfitting.
Automatic methods suffer from the degradation of the detection if the previous stage fails to detect the target.
For instance, the automatic techniques need to segment players on the court to record their current position.
The program requires customization to ensure that it can only detect two players but not the referees and fans.
On the flip side, our labeling process reduces labeling errors and maintains annotation accuracy based on the detection process.

\section{Annotation Process}
\label{annotation-process}

%% introduce S^2 and the number of annotators (6)
The annotation process was carried out mainly with the badminton experts using the shot-by-shot (S\textsuperscript{2}) labeling tool \cite{DBLP:conf/apnoms/HuangHLIW22}.
It is a microscopic labeling system for badminton and has been demonstrated as having efficient and effective labeling quality by badminton experts.
For example, the S\textsuperscript{2} labeling tool has proven to meet the requirement of completing the labeling process on match day for tactic investigation for the next match.
The collection process of stroke-level annotation of a match using the S\textsuperscript{2} labeling tool can be mainly divided into four stages:
\begin{itemize}
    \item Match preparation: The collected matches are described in Section \ref{shuttleset-dataset}.
    To provide both practical and efficient requirements, the frame rate of the video is 30 frames per second and the resolution is 1280x720 (720p).
    Afterwards, the metadata of the match is input, e.g., the tournament name, the tournament level, round in the tournament, date, player names, played sets, duration (in minutes), and the four corners of the court boundary lines are marked for computing the corresponding homography matrix.
    \item Rally Segmentation: Annotators are required to segment the start and end of each rally and the scoring result by recording the time of the corresponding event, which is supported by the shuttle trajectory to reduce the switching time of annotators.
    \item Video preprocessing: S\textsuperscript{2} labeling tool splits the match video into multiple rally videos to avoid heavily wasting computing resources and to eliminate execution time due to the break time.
    Then, the video is preprocessed with detection techniques such as trajectory detection, player detection, and skeleton detection.
    For the use of labeling, only trajectory detection is employed.
    \item Shot-by-shot labeling: Initiate the labeling process. Annotators need to click on the screen and select the corresponding shot types with a one-pass labeling operation. The location information can be labeled by directly clicking on the video screen to avoid manual alignment errors, and the corresponding clicking point is displayed simultaneously on the video screen for instant confirmation. The details will be introduced next.
\end{itemize}

\begin{figure}
    \centering
    \includegraphics[width=\linewidth]{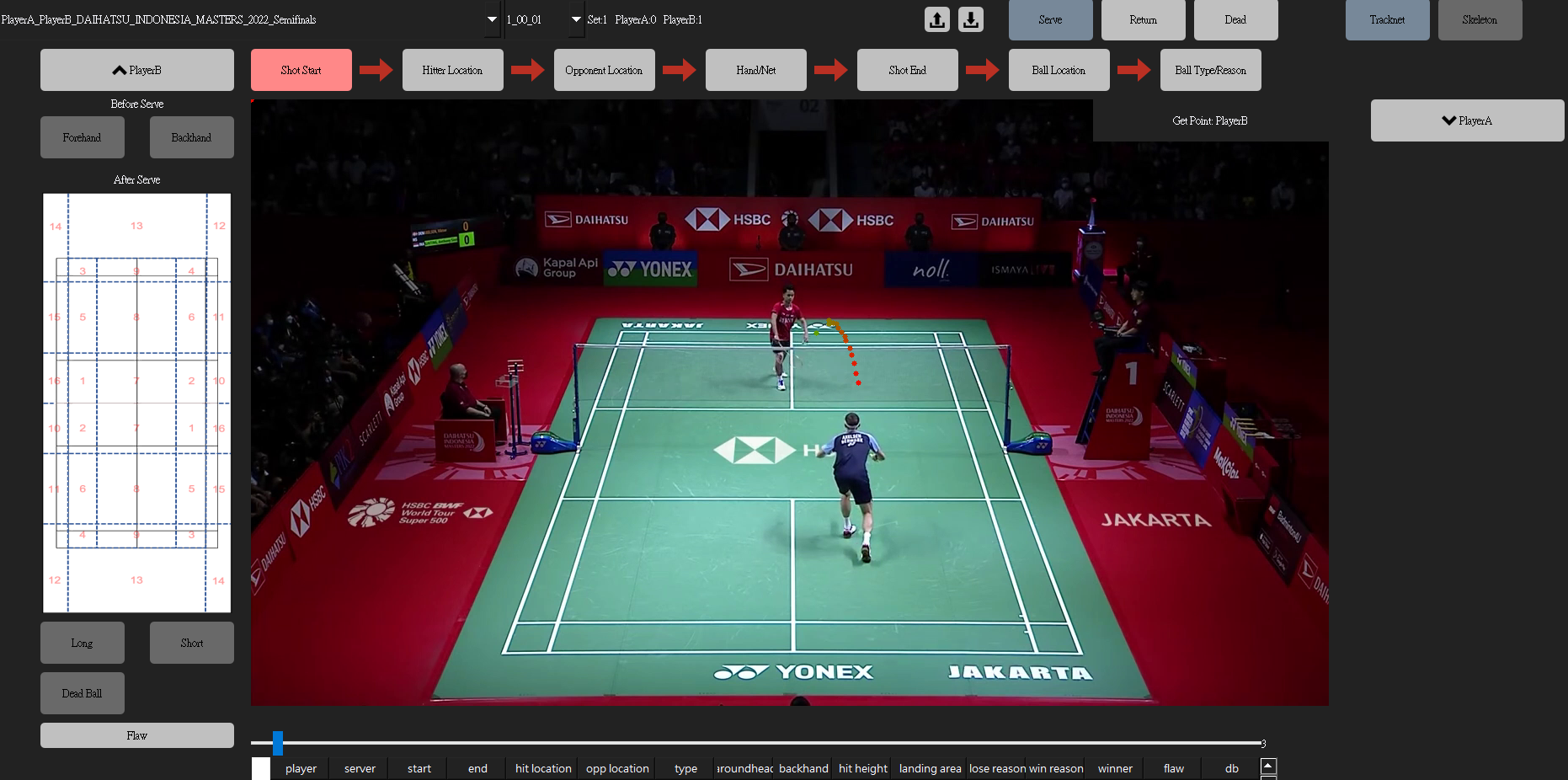}
    \caption{An interface of the labeling tool with seven steps.}
    \label{ui-example}
\end{figure}

\begin{figure}
    \centering
    \includegraphics[width=\linewidth]{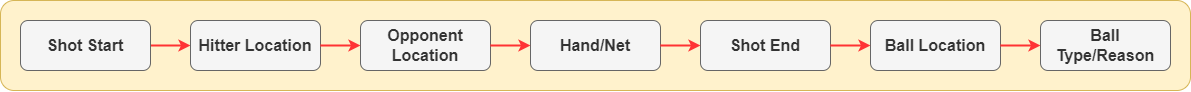}
    \caption{The labeling process of the fourth stage. The corresponding mark turns to red to remind the current step.}
    \label{zoom-process}
\end{figure}

%% stage 2: introduce the labeling process
A group of six dedicated annotators with badminton knowledge was assembled and supervised for ShuttleSet annotation with the S\textsuperscript{2} labeling tool.
Figure \ref{ui-example} shows an interface of the labeling process of stage four and Figure \ref{zoom-process} illustrates the labeling process consisting of seven steps, which is optimized to achieve a one-pass labeling operation.
The labeling of each rally starts from the time of the serve type to the end of the dead bird type, which requires labeling in order: 1) the hitting time, 2) the location of the player returning the stroke, 3) the location of the opponent, 4) if the shuttle is hit with a backhand, around the head, and above or below the net, 5) the returning time of the opponent (the landing time of a shuttle if the rally ends), 6) the landing location of a shuttle, and 7) the corresponding shot type.
The labeling cycle is then repeated for each shot in a rally, where the next shot starts from the opponent's returning time of the previous returning time and is treated as the hitting time of the current shot until the dead bird happens as the last stroke.

The labeled records are illustrated below in the labeling interface following the format in Section \ref{shuttleset-dataset}, and it can be instantaneously saved by clicking the uploaded button.
To reduce the labeling error, the S\textsuperscript{2} labeling tool also incorporates a fool-proof design to avoid unintended manual clicking errors or misjudgment.
In addition, rallies in a match will be randomly sampled for quality checking to verify the correctness of the stroke-level labels.
\section{Benchmarks}
\label{experiments}

% goal of the dataset
The primary goal of ShuttleSet is to obtain high-quality ML and DL models capable of analyzing players' patterns on a wide variety of tactical strokes.
Since ShuttleSet is not limited to the specific use as most of the standard public datasets, the train, validation, and test sets are able to customize based on the corresponding tasks instead of being pre-defined.
As such, we will relate to present various application tasks in this paper and apply state-of-the-art methods to address these problems.
They can be used as benchmarks for researchers who aim to advance their approaches on turn-based sequences following the corresponding train, validation, and test sets.

% what will we present in the following sections
In this section, we present 3 application scenarios related to the performance on ShuttleSet, including measuring the win probability of each stroke in a rally (shot influence \cite{DBLP:conf/icdm/WangCYWFP21,10.1145/3551391}, Section \ref{shot-influence}), forecasting the future strokes (stroke forecasting \cite{DBLP:conf/aaai/WangSCP22}, Section \ref{stroke-forecasting}) and the movements of both players (movement forecasting \cite{DBLP:journals/corr/abs-2211-12217}, Section \ref{movement-forecasting}) given the previous strokes.
Following their evaluation settings for data separation and evaluation metrics, the shot influence task formulates the task as predicting the final outcome of a rally and adopts the latest 10 matches as the test set and the remaining for the train set to evaluate model effectiveness for inferring the win probability of each stroke in the unseen matches.
The quality of a predicted probability is measured with the area under the receiver operator curve (AUC), accuracy (ACC), and brier score (BR).
The coordinates of both shuttle and players' movements are preprocessed with standardization.

Both stroke forecasting and movement forecasting utilize the first 80\% of rallies of each match as the train set and the remaining 20\% of rallies as the test set to enforce the model equipped with the information of previous matches of each player.
As these forecasting tasks exist stochasticity, both models are required to generate 10 sequences for each rally, and the sequence closest to the ground truth is the final prediction for evaluation.
In addition, three different lengths of given strokes (i.e., given 8, 4, and 2 of each rally) are used to examine the robustness of models in terms of the variable encoding strokes.
The evaluation metrics include cross entropy (CE) for shot type evaluation, mean absolute error (MAE), and mean square error (MSE) for location evaluation of both shuttles and two players.

\subsection{Benchmark 1: Shot Influence}
\label{shot-influence}

\citet{DBLP:conf/icdm/WangCYWFP21,10.1145/3551391} introduces the shot influence task by framing as predicting which player wins the rally by giving the stroke-level information (e.g., shot type, player location, if returning around the head), which can be used to measure the quality of each shot in a rally for enhancing players' decision-making confidence.
Generally, the state-of-the-art model is ShuttleScorer \cite{10.1145/3551391}, which consists of a shot encoder for encoding stroke-level information with various timestamps, a pattern extractor for considering short-term patterns alternatively, and a rally encoder with a Transformer encoder \cite{DBLP:conf/nips/VaswaniSPUJGKP17}.
We also implemented a bi-directional gated recurrent unit (Bi-GRU) \cite{DBLP:conf/ssst/ChoMBB14}, Prototype Sequence Network (ProSeNet) \cite{DBLP:conf/kdd/MingXQR19}, DeepMoji \cite{DBLP:conf/emnlp/FelboMSRL17}, Ordered neurons LSTM (ON-LSTM) \cite{DBLP:conf/iclr/ShenTSC19}, Transformer \cite{DBLP:conf/nips/VaswaniSPUJGKP17}, Reformer \cite{DBLP:conf/iclr/KitaevKL20}, and TokenLearner \cite{DBLP:conf/nips/RyooPADA21} as the baselines for comparison.

\begin{table}
    \small
    \centering
    \caption{Performance of the shot influence in terms of AUC, accuracy (ACC), and brier score (BR).}
    \begin{tabular}{cc|ccc}
    \toprule
    Task & Model & AUC ($\uparrow$) & ACC ($\uparrow$) & BR ($\downarrow$) \\
    \midrule
    \multirow{8}{*}{Shot Influence} & 
    ProSeNet & 0.5689 & 0.5000 & 0.2500 \\
    & Bi-GRU & 0.6603 & 0.7054 & 0.2158 \\
    & ON-LSTM & 0.6766 & 0.6808 & 0.2244 \\
    & DeepMoji & 0.6786 & 0.6802 & 0.2238 \\
    & Transformer & 0.6846 & 0.7570 & 0.1986 \\
    & Reformer & 0.7046 & 0.7454 & 0.1994 \\
    & TokenLearner & 0.7166 & 0.7698 & 0.1933 \\
    & ShuttleScorer & \textbf{0.8371} & \textbf{0.7869} & \textbf{0.1574} \\
    \bottomrule
\end{tabular}
    \label{tab:shot-influence}
\end{table}

Table \ref{tab:shot-influence} shows the baseline experiments of ShuttleScorer in the average of 5 random seeds.
The hyper-parameters are identical to the implementation details in the paper.
As one can observe, Bi-GRU only reaches 0.6603, 0.7054, and 0.2158 in terms of AUC, ACC, and BR, respectively.
In addition, the performance of AUC, accuracy, and brier score of ShuttleScorer significantly outperform Bi-GRU for 26.8\%, 11.6\%, and 27.1\%, respectively, but there still remains a gap to be improved since the current AUC score reaches 0.8371.
This gives a good indication that ShuttleSet provides more diverse matches and poses a worthwhile challenge for the research community to design turn-based approaches.
Moreover, this also provides an intuition into the technique of incorporating players with only a few matches as shown in Figures \ref{men-matrix} and \ref{women-matrix}.
It is interesting to see how future work tackles these imbalanced and limited players.

\subsection{Benchmark 2: Stroke Forecasting}
\label{stroke-forecasting}

\begin{table*}
    \small
    \centering
    \caption{Performance of the stroke forecasting for different observed stroke sequence lengths ($\tau$).}
    \begin{tabular}{c|c|ccc|ccc|cccccc}
    \toprule
    & & \multicolumn{3}{c|}{$\tau=8$} & \multicolumn{3}{c|}{$\tau=4$} & \multicolumn{3}{c}{$\tau=2$} \\
    \cmidrule{3-11}
    Task & Model & CE & MSE & MAE & CE & MSE & MAE & CE & MSE & MAE \\
    \midrule
        \multirow{5}{*}{Stroke Forecasting} & Seq2Seq & 3.0588 & 1.8414 & 1.4840 & 3.2044 & 1.7839 & 1.4534 & 3.1116 & 1.7348 & 1.4404 \\
        & CFLSTM & 2.8427 & 2.3835 & 1.7551 & 2.8979 & 2.4119 & 2.7676 & 2.8914 & 2.3348 & 1.7173 \\
        & TF & 2.9555 & 1.9530 & 1.5289 & 3.0254 & 1.7790 & 1.4623 & 2.9477 & 1.8207 & 1.4667 \\
        & dNRI & 2.9736 & 2.8238 & 1.9245 & 2.9689 & 2.6518 & 1.8691 & 2.9819 & 2.9194 & 1.9540 \\
        & DMA-Nets & 3.1041 & 2.2182 & 1.6494 & 2.9932 & 2.1020 & 1.5940 & 2.9600 & 2.0101 & 1.5565 \\
        & ShuttleNet & \textbf{2.4125} & \textbf{1.8121} & \textbf{1.3582} & \textbf{2.4065} & \textbf{1.5673} & \textbf{1.3777} & \textbf{2.3901} & \textbf{1.5034} & \textbf{1.3615} \\
    \bottomrule
\end{tabular}
    \label{tab:stroke-forecasting}
\end{table*}

\begin{table*}
    \small
    \centering
    \caption{Performance of the movement forecasting for different observed stroke sequence lengths ($\tau$).}
    \begin{tabular}{c|c|ccc|ccc|cccccc}
    \toprule
    & & \multicolumn{3}{c|}{$\tau=8$} & \multicolumn{3}{c|}{$\tau=4$} & \multicolumn{3}{c}{$\tau=2$} \\
    \cmidrule{3-11}
    Task & Model & CE & MSE & MAE & CE & MSE & MAE & CE & MSE & MAE \\
    \midrule
        \multirow{7}{*}{Movement Forecasting} 
        & Seq2Seq & 2.3819 & 1.1575 & 1.6706 & 2.3971 & 1.1279 & 1.6556 & 2.4280 & 1.1463 & 1.6588 \\
        & TF & 2.3969 & 1.1716 & 1.6772 & 2.4102 & 1.1580 & 1.6681 & 2.4551 & 1.1400 & 1.6559 \\
        & ShuttleNet & 2.3801 & 1.1926 & 1.6727 & 2.3810 & 1.1312 & 1.6255 & 2.3655 & \textbf{1.0832} & \textbf{1.5852} \\
        & dNRI & 2.9587 & 1.5351 & 1.6758 & 2.9626 & 1.6084 & 1.7219 & 2.9526 & 1.5313 & 1.6614 \\
        & GCN\textsubscript{PM} & 2.3910 & 1.3145 & 1.7969 & 2.4027 & 1.2954 & 1.7703 & 2.3641 & 1.2535 & 1.7372 \\
        & R-GCN\textsubscript{PM} & 2.3888 & 1.1128 & 1.6165 & 2.5110 & 1.1971 & 1.6571 & 2.4370 & 1.2636 & 1.6873 \\
        & DyMF & \textbf{2.3566} & \textbf{1.0963} & \textbf{1.5875} & \textbf{2.3641} & \textbf{1.1107} & \textbf{1.6107} & \textbf{2.3146} & 1.1371 & 1.6142 \\
    \bottomrule
\end{tabular}
    \label{tab:movement-forecasting}
\end{table*}

In addition to the classification task, we also implement forecasting tasks using ShuttleSet, which is generally more challenging than the classification task.
\citet{DBLP:conf/aaai/WangSCP22} proposes the stroke forecasting task, which is beneficial for not only simulating players' tactics but also assessing the returning probability of future strokes for storytelling.
This task is defined as predicting future strokes including shot types and the corresponding destination locations based on the previous strokes in a rally.
We reproduced the state-of-the-art model, ShuttleNet \cite{DBLP:conf/aaai/WangSCP22}, and the baselines Seq2Seq \cite{DBLP:conf/nips/SutskeverVL14}, CF-LSTM \cite{DBLP:conf/aaai/XuYD20}, TF \cite{DBLP:conf/icpr/GiuliariHCG20}, dNRI \cite{Graber_2020_CVPR}, and DMA-Nets \cite{DBLP:conf/aaai/JiSFCRL21} for illustrating the performance of shot types and landing locations.
Seq2Seq uses two long short-term memory (LSTM) \cite{DBLP:journals/corr/SakSB14} as an encoder and a decoder.
ShuttleNet contains a Transformer-based player extractor for modeling styles of each player separately, a Transformer-based rally extractor for capturing the current progress of a rally, and a position-aware gated fusion network for fusing contexts of two players and rally based on the positions of the rally.
All hyper-parameters are the same as the original paper.

Table \ref{tab:stroke-forecasting} reports the performance in different given stroke sequences, which can be seen that ShuttleNet consistently outperforms the Seq2Seq model in terms of both shot type prediction (CE) and location prediction (MSE and MAE).
Furthermore, the shot type performance of ShuttleNet does not change significantly across various encoding strokes, while the performance of Seq2Seq is more unstable.
It is worth noting that both models degrade the shot type performance compared with those in the original paper, which is because there are 18 shot types while the original paper only used 10 shot types for simplification.
This result points out that the fine-grained definitions of shot types in ShuttleSet increase the uncertainty of current models, which is expected to close the gap from the research community.

\subsection{Benchmark 3: Movement Forecasting}
\label{movement-forecasting}

As stroke forecasting only tackles the future shot types and the corresponding locations, \citet{DBLP:journals/corr/abs-2211-12217} introduces a more challenging task: movement forecasting, where the goal is to predict the same targets of stroke forecasting as well as the movements of players.
This task makes the analyzing scenarios more complete since coaches and players can take advantage of it for investigating the physical ability and the habit of defensive positions of players.
DyMF \cite{DBLP:journals/corr/abs-2211-12217} is the state-of-the-art graph-based model for the movement forecasting task, which represents player interactions in a rally as a strategic graph.
DyMF consists of the interaction style extractor for dynamically capturing the behaviors of mutual interactions and players' tactics based on previous patterns, and the hierarchical fusion for fusing the style of both players with their opponents' styles and rally interactions.
We reproduce DyMF, Seq2Seq, TF \cite{DBLP:conf/icpr/GiuliariHCG20}, ShuttleNet \cite{DBLP:conf/aaai/WangSCP22}, dNRI \cite{Graber_2020_CVPR}, GCN\textsubscript{PM} \cite{DBLP:conf/iclr/KipfW17}, and R-GCN\textsubscript{PM} \cite{DBLP:conf/esws/SchlichtkrullKB18} as the baselines and all hyper-parameters are the same as in the original paper.

Table \ref{tab:movement-forecasting} illustrates the performance of Seq2Seq and DyMF in different observed stroke sequences.
DyMF consistently outperforms Seq2Seq for shot type and movement performance, which further reaches lower uncertainty even for a more challenging task compared with ShuttleNet in the stroke forecasting task.
It is interesting that Seq2Seq performs stably for shot type performance in this task, while DyMF does not have a pattern of different lengths of given strokes.
This benchmark provides intuitions and raises an opportunity for the graph-based models that are able to be utilized in the turn-based matches instead of only using sequence-based models due to the constraint of heterogeneous relations of players themselves and between players.

\section{Deployment: Visualization Platform}
\label{visualization-platform}

\subsection{Deployment Setup}
As the records in ShuttleSet are labeled from real-world top-ranking matches, in addition to providing benchmarks for advancing turn-based methods, we also construct a visualization platform to deploy these records for analysis.
This benefits the direct usage and reduces the technical gaps for the sports community, e.g., coaching players and tactic investigations in data-driven methods.

Specifically, we have built an analysis platform\footnote{https://coachai.cs.nctu.edu.tw:55000} to visualize statistics from ShuttleSet with respect to various aspects for coaches and players to review instateneously without repeating match videos.
Records in ShuttleSet are converted to the JSON format for the Flask engine\footnote{https://flask.palletsprojects.com/en/2.2.x/} to consume.
The platform consists of two major components, analysis for each match and summarized comparisons from all matches of the same matchup.
Here we illustrate the analysis for each match as an example, which provides position analysis, win/lose analysis \cite{Wang2020badminton}, and shot types analysis.

\subsection{Case Studies}
\begin{figure*}
    \centering
    \includegraphics[width=\linewidth]{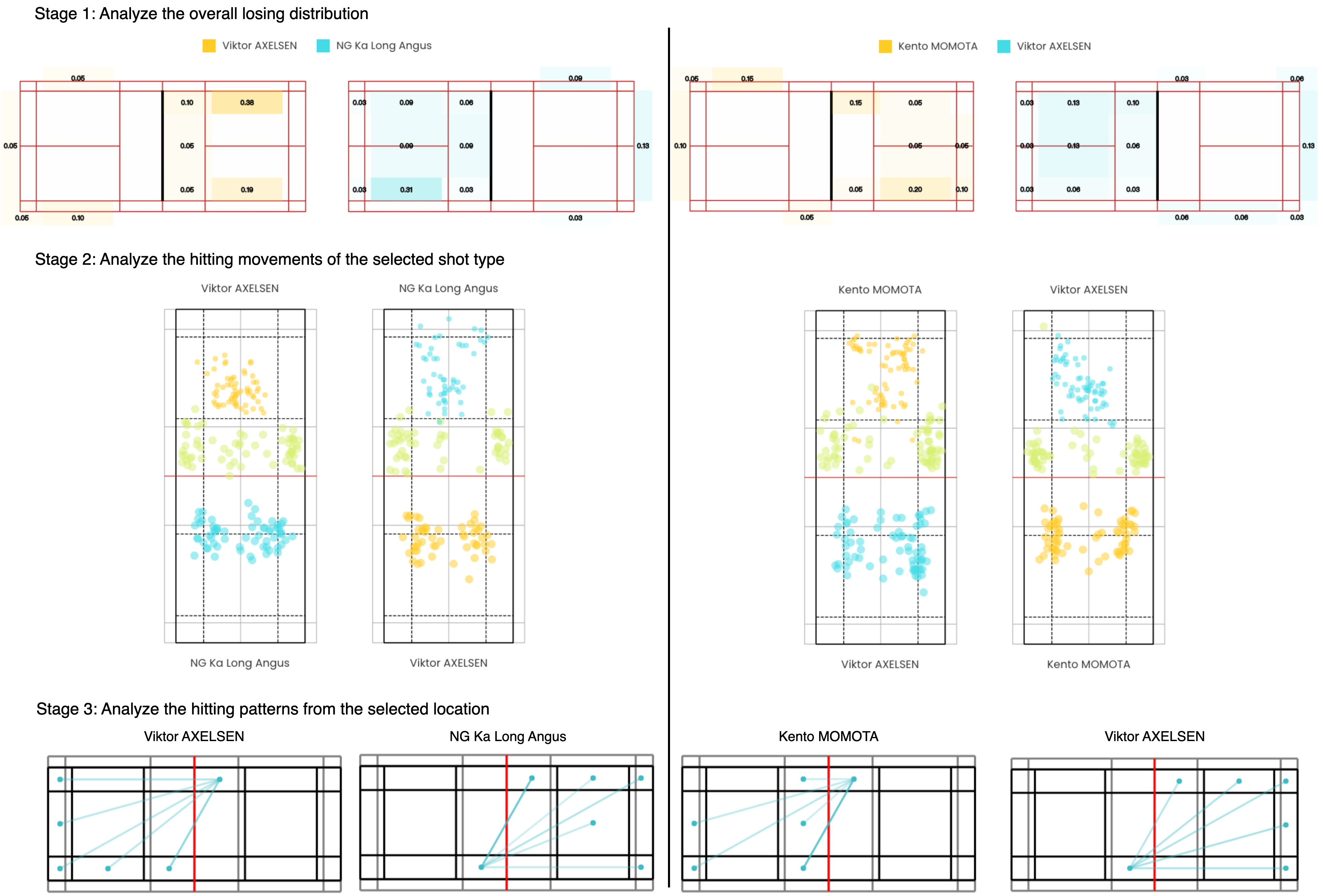}
    \caption{Three analysis techniques are presented in the Semi-finals (Viktor Axelsen vs. Ng Ka Long Angus) and Finals (Kento Momota vs. Viktor Axelsen) of the Malaysia Masters 2020. For each match, the winning player is displayed on the left side and the losing player is on the right side.}
    \label{platform}
\end{figure*}

Figure \ref{platform} illustrates three analysis techniques with Finals (Kento Momota vs. Viktor Axelsen) and Semi-finals (Viktor Axelsen vs. Ng Ka Long Angus) in the Men's Singles of the Malaysia Masters 2020.
The first stage shows the overall losing distribution of each player separated by the grid system according to Figure \ref{court}, where the area with a higher proportion is colored with lower transparency and vice versa.
The second stage demonstrates the players' movements when each player returns the net shot.
The blue and yellow points on the upper court represent the locations of the player when the opponent returns the stroke (i.e., the last stroke of the net shot), and the yellow and blue points on the lower court represent the locations of the opponent when the opponent returns the stroke.
The green points represent the locations of the player when returning the net shot.
The third stage illustrates the analysis of the hitting patterns from the selected location (grid 1 in the figure), which demonstrates the five most common returning locations.

In the first stage, we can observe that the right side of Ng Ka Long Angus is the main weak area, which is also true for Viktor Axelsen, while during the Finals, both Kento Momota and Viktor Axelsen have more uniform losing distributions.
This shows that it was hard for Ng Ka Long Angus to get the winning point if he attacked to the middle of the court, while Kento Momota was able to win the rally by returning to the middle court.
In addition, the outside distributions of Viktor Axelsen with Kento Momota are higher than those with Ng Ka Long Angus (0.25 vs 0.37), which indicates that Kento Momota is able to return strokes that cause Viktor Axelsen to make unforced errors (i.e., return to outside by himself).
As Kento Momota is a left-handed player, we can see that the front court with a backhand (grid 1) is the main area where he will lose a point in the front court, which suggests the opponent has an opportunity to consider the strategy when returning to the front court.

To delve into the strategies of the net shot in the front court, the second stage illustrates the hitting movements of both players for dealing with the net shot.
From the Semi-finals, it is obvious that the positions of Viktor Axelsen mainly stand in the middle of the court when preparing to return the net shot, while the positions of Ng Ka Long Angus show some patterns in the back court.
This analysis indicates that Ng Ka Long Angus is often mobilized by Viktor Axelsen since he needs to move a greater distance from the back court to the front court, which consumes more physical energy; however, this issue does not occur for Viktor Axelsen as he is able to move to the center of the court rapidly after he returns the stroke.
Nonetheless, this situation does not hold during the Finals, as Kento Momota and Viktor Axelsen fail to stand in the middle of the court consistently to prepare for the net shot, while Kento Momota stands nearer (yellow points in the right subfigure) the net when Viktor Axelsen returns the net shot.
Since Kento Momota is proficient in performing net shots \cite{kento_momota_study}, his opponents often find it difficult to get the point from returning to the front court, which might be one of the reasons why Kento Momota won the championship in the Malaysia Masters 2020.

In the third stage, we aim to further investigate the common returning patterns from grid 1.
We can see that Viktor Axelsen has the same common returning areas when in the semi-finals and finals, while Ng Ka Long Angus is more likely to return to the right side of his opponent.
It is interesting that Kento Momota is likely to return to the front court, which also verifies the confident skills of net shots as the aforementioned advantage.
\section{Discussion}
Throughout this paper, we claim that ShuttleSet's wider variety of stroke-level records is able to foster more advanced ML and DL models for mitigating the gap between the research and sports communities.
Although ShuttleSet is the largest public turn-based sports dataset with microscopic information, we illustrate several benchmarks with both classification tasks (Benchmark 1) and forecasting tasks (Benchmarks 2 and 3) and provide some intuitions and insights into the current state-of-the-art approaches.
From the benchmarks, the performance looks decent but still requires improvements to produce more robust and better results for these tasks, which also provide opportunities for both sequence- and graph-based approaches to tackle the turn-based sequences.
It is also expected that there will be other beneficial applications that are proposed and investigated for turn-based sports analytics, e.g., style comparison and limited records of players.
Moreover, we demonstrate the practical deployment usage of ShuttleSet with a visualization platform using two matches as paradigms, which enables the sports community to investigate players' tactics from previous matches without repeating the videos.
The platform is designed with a friendly user interface and does not require any technical background, which can be used directly.
\section{Conclusion}
In this paper, we present the ShuttleSet dataset, which provides both sports analysis and academic research communities with the largest and challenging badminton dataset with stroke-level records to stimulate and improve novel ML and DL methods on turn-based sequences.
Instead of detecting automatically, ShuttleSet was created by human annotation to obtain not only reliable and structural ground truth from matches to strokes but also a wide variety of general match videos of high-ranking men's and women's singles.
In addition, several statistics illustrate the compositions and intuitions of ShuttleSet to mitigate the gaps for researchers not familiar with turn-based sports.
Based on the dataset, multiple paradigms from the experiments have showcased the practical usage of turn-based analytics including an online visualization platform for badminton coaches to investigate players' tactics and patterns.
These applications represent the benchmarks for the research community to advance their methods as well as the inspirations for the sports community to foster multifaceted analysis applications.
To date, there is still a significant gap between the sports community and the research community on turn-based research applications, and so we hope this work will inspire both communities to advance techniques for closing that gap.

\section*{Acknowledgments}
This work was supported by the Ministry of Science and Technology of Taiwan under Grants 110-2221-E-A49-063-MY3, 111-2622-E-A49-009, 111-3114-H-A49-001, and 112-2425-H-A49-001.

\bibliographystyle{ACM-Reference-Format}
\balance
\bibliography{reference}

% \clearpage
\appendix
\section{Appendix}

\begin{table*}
    \small
    \centering
    \caption{Detailed information of each labeled match.}
    \begin{tabular}{ccccccc}
    \toprule
    Tournament & Round & Year & Set & Duration (m) & Winner & Loser \\
    \midrule
    Fuzhou Open 2018 & Finals & 2018 & 3 & 67 & Kento Momota & Chou Tien Chen \\
    Denmark Open 2018 & Finals & 2018 & 3 & 77 & Kento Momota & Chou Tien Chen \\
    Malaysia Open 2018 & Quarter-Finals & 2018 & 2 & 53 & Kento Momota & Chou Tien Chen \\
    Fuzhou Open 2019 & Finals & 2019 & 3 & 83 & Kento Momota & Chou Tien Chen \\
    World Tour Finals & Group Stage & 2019 & 2 & 52 & Chen Long & Chou Tien Chen \\
    KOREA OPEN 2019 & Finals & 2019 & 2 & 53 & Kento Momota & Chou Tien Chen \\
    Denmark Open 2019 & Quarter-Finals & 2019 & 2 & 54 & Chen Long & Chou Tien Chen \\
    Fuzhou Open 2019 & Semi-Finals & 2019 & 2 & 45 & Chou Tien Chen & Anders Antonsen \\
    Sudirman Cup 2019 & Quarter-Finals & 2019 & 2 & 36 & Chou Tien Chen & Jonatan Christie \\
    Sudirman Cup 2019 & Group Stage & 2019 & 2 & 41 & Chou Tien Chen & Ng Ka Long Angus \\
    Indonesia Open 2019 & Quarter-Finals & 2019 & 3 & 76 & Chou Tien Chen & Jonatan Christie \\
    Thailand Masters 2020 & Semi-Finals & 2020 & 2 & 37 & Ng Ka Long Angus & Shi Yu Qi \\
    Malaysia Masters 2020 & Semi-Finals & 2020 & 2 & 44 & Viktor Axelsen & Ng Ka Long Angus \\
    Malaysia Masters 2020 & Quarter-Finals & 2020 & 3 & 65 & Ng Ka Long Angus & Jonatan Christie \\
    Malaysia Masters 2020 & Quarter-Finals & 2020 & 3 & 68 & Viktor Axelsen & Chen Long \\
    Malaysia Masters 2020 & Finals & 2020 & 2 & 54 & Kento Momota & Viktor Axelsen \\
    Indonesia Masters 2020 & Quarter-Finals & 2020 & 3 & 68 & Anders Antonsen & Jonatan Christie \\
    Indonesia Masters 2020 & Finals & 2020 & 3 & 71 & Anthony Sinisuka Ginting & Anders Antonsen \\
    Indonesia Masters 2020 & Semi-Finals & 2020 & 2 & 43 & Anthony Sinisuka Ginting & Viktor Axelsen \\
    All England Open 2020 & Quarter-Finals & 2020 & 2 & 40 & Viktor Axelsen & Shi Yu Qi \\
    HSBC BWF World Tour Finals 2020 & Quarter-Finals & 2020 & 3 & 55 & An Se Young & Carolina Marin \\
    HSBC BWF World Tour Finals 2020 & Quarter-Finals & 2020 & 3 & 50 & Anthony Sinisuka Ginting & Lee Zii Jia \\
    HSBC BWF World Tour Finals 2020 & Quarter-Finals & 2020 & 2 & 34 & Evgeniya Kosetskaya & Michelle Li \\
    HSBC BWF World Tour Finals 2020 & Quarter-Finals & 2020 & 3 & 65 & Ng Ka Long Angus & Kidambi Srikanth \\
    HSBC BWF World Tour Finals 2020 & Quarter-Finals & 2020 & 2 & 42 & Pusarla V. Sindhu & Pornpawee Chochuwong \\
    HSBC BWF World Tour Finals 2020 & Semi-Finals & 2020 & 2 & 40 & Carolina Marin & Pornpawee Chochuwong \\
    HSBC BWF World Tour Finals 2020 & Semi-Finals & 2020 & 3 & 60 & Anders Antonsen & Viktor Axelsen \\
    Yonex Thailand Open 2021 & Quarter-Finals & 2021 & 2 & 44 & An Se Young & Ratchanok Intanon \\
    Yonex Thailand Open 2021 & Quarter-Finals & 2021 & 2 & 45 & Mia Blichfeldt & Busanan Ongbamrungphan \\
    Yonex Thailand Open 2021 & Quarter-Finals & 2021 & 2 & 44 & Ng Ka Long Angus & Lee Cheuk Yiu \\
    Yonex Thailand Open 2021 & Quarter-Finals & 2021 & 3 & 66 & Anthony Sinisuka Ginting & Rasmus Gemke \\
    Yonex Thailand Open 2021 & Quarter-FInals & 2021 & 2 & 44 & Carolina Marin & Supanida Katethong \\
    Yonex Thailand Open 2021 & Quarter-Finals & 2021 & 2 & 40 & Viktor Axelsen & Jonatan Christie \\
    Yonex Thailand Open 2021 & Semi-Finals & 2021 & 3 & 63 & Viktor Axelsen & Anthony Sinisuka Ginting \\
    Yonex Thailand Open 2021 & Finals & 2021 & 2 & 44 & Viktor Axelsen & Ng Ka Long Angus \\
    Toyota Thailand Open 2021 & Quarter-Finals & 2021 & 2 & 44 & An Se Young & Pornpawee Chochuwong \\
    Toyota Thailand Open 2021 & Quarter-Finals & 2021 & 3 & 81 & Anders Antonsen & Sameer Verma \\
    Toyota Thailand Open 2021 & Quarter-Finals & 2021 & 2 & 34 & Carolina Marin & Neslihan Yigit \\
    Toyota Thailand Open 2021 & Quarter-Finals & 2021 & 3 & 77 & Hans-Kristian Solberg Vittinghus & Lee Cheuk Yiu \\
    Toyota Thailand Open 2021 & Quarter-Finals & 2021 & 2 & 41 & Viktor Axelsen & Luiw Daren \\
    Toyota Thailand Open 2021 & Quarter-Finals & 2021 & 2 & 38 & Ratchanok Intanon & Pusarla V. Sindhu \\
    Toyota Thailand Open 2021 & Semi-Finals & 2021 & 2 & 51 & Carolina Marin & An Se Young \\
    Toyota Thailand Open 2021 & Semi-Finals & 2021 & 2 & 45 & Hans-Kristian Solberg Vittinghus & Anders Antonsen \\
    Toyota Thailand Open 2021 & Finals & 2021 & 2 & 40 & Viktor Axelsen & Hans-Kristian Solberg Vittinghus \\
    \bottomrule
\end{tabular}
    \label{tab:labeled-match}
\end{table*}

\subsection{Details of labeled matches.}
\label{detail-match}
Table \ref{tab:labeled-match} presents the detailed information of each labeled match, including the tournament name, round, year, played set, duration, winner, and loser.
Generally, the matches were collected from 2018 to 2021 with high-ranking players.
% The sampled dataset, detailed match information, and homography matrices in ShuttleSet are provided in the link\footnote{shorturl.at/quAVY} and the full stroke-level dataset will be released in the GitHub repo in the camera-ready version.

\end{document}